\documentclass{article}
\usepackage{dsfont}
\usepackage{cite}
\usepackage{amsmath,amssymb,amsfonts}
\usepackage{algorithmic}
\usepackage{graphicx}
\usepackage{textcomp}
\usepackage{xcolor}
\usepackage{comment}
\usepackage{multirow}
\usepackage{spconf}
\usepackage[singlelinecheck=false]{caption}
\usepackage{url}
\usepackage{tikz}
\usetikzlibrary{positioning}
\usepackage[linesnumbered,lined,boxed,commentsnumbered,ruled]{algorithm2e}
\usepackage{array}
\usepackage{color,soul}
\usepackage{pifont}
\def\cmark{\ding{51}} 
\def\xmark{\ding{55}}
\usepackage{makecell}
\usepackage{array}
\usepackage{units}
\usepackage{multirow}
\usepackage{booktabs}
\usepackage{makecell}
\usepackage{tabularx}
\usepackage{url}
\usepackage{tikz}
\usepackage{pgfplots}
\usepgfplotslibrary{colorbrewer}
\pgfplotsset{cycle list/Set1-9}
\tikzset{every picture/.style={line width=1pt}}
\usetikzlibrary{patterns}
\usepgfplotslibrary{fillbetween}
\pgfplotsset{
 tick label style = {font=\sansmath\sffamily\scriptsize},
 every axis label = {font=\sansmath\sffamily\scriptsize},
 y label style={at={(0.05,0.5)}},
 legend style = {font=\sansmath\sffamily\scriptsize},
 label style = {font=\sansmath\sffamily\scriptsize},
}
\pgfplotsset{compat=1.3} %
\DeclareMathOperator*{\argmax}{\arg\!\max}
\DeclareMathOperator*{\argmin}{\arg\!\min}

\title{Deep Metric Learning-Based Semi-Supervised Regression\\With Alternate Learning}
\name{Adina Zell, Gencer Sumbul, Beg\"{u}m Demir}
\address{Faculty of Electrical Engineering and Computer Science, Technische Universit{\"a}t Berlin, Germany}
\global\hyphenpenalty=1
\begin{document}
\maketitle

\begin{abstract}
This paper introduces a novel deep metric learning-based semi-supervised regression (DML-S2R) method for parameter estimation problems. The proposed DML-S2R method aims to mitigate the problems of insufficient amount of labeled samples without collecting any additional sample with a target value. To this end, it is made up of two main steps: i) pairwise similarity modeling with scarce labeled data; and ii) triplet-based metric learning with abundant unlabeled data. The first step aims to model pairwise sample similarities by using a small number of labeled samples. This is achieved by estimating the target value differences of labeled samples with a Siamese neural network (SNN). The second step aims to learn a triplet-based metric space (in which similar samples are close to each other and dissimilar samples are far apart from each other) when the number of labeled samples is insufficient. This is achieved by employing the SNN of the first step for triplet-based deep metric learning that exploits not only labeled samples but also unlabeled samples. For the end-to-end training of DML-S2R, we investigate an alternate learning strategy for the two steps. Due to this strategy, the encoded information in each step becomes a guidance for learning phase of the other step. The experimental results confirm the success of DML-S2R compared to the state-of-the-art semi-supervised regression methods. The code of the proposed method is publicly available at \url{https://git.tu-berlin.de/rsim/DML-S2R}.
\end{abstract}

\section{Introduction}
The accurate estimation of parameters from specific data (e.g., estimation of carbon monoxide concentration in the environment, estimation of forest parameters, estimation of glucose concentration in diabetic patients, etc.) is an important research field in machine learning \cite{Demir2014AMC}. The use of regression methods, which aim at learning functional relations between a set of variables and corresponding target values, is an effective way for parameter estimation problems. Accordingly, several regression methods (e.g., random forest regression~\cite{Breiman:2001}, support vector regression~\cite{Demir2014AMC}, deep learning-based regression~\cite{DeepPose, FacialPointDetection}, etc.) have been introduced in the literature. The success of these methods depends on the quantity and quality of training samples for which the respective target values are available. The quantity of training samples depends on the number of available labeled samples, while the quality of training samples depends on their capability to represent the real sample distribution. A small amount of labeled samples with insufficient quality can lead to inadequate modeling of the regression task. However, collecting labeled samples is often expensive and complex as producing reference measures may require significant time \cite{Demir2014AMC}. To address this issue, semi-supervised regression (SSR) methods, which aim at jointly using the information of both labeled and unlabeled samples in the learning phase of the regression algorithm, have been introduced in \cite{CoTraining, SSLRegression, TIMILSINA2021107188}.

In recent years, deep metric learning (DML) has been utilized in the framework of SSR. DML operates on sample tuples (e.g., triplets) to learn a metric space, in which similar samples are mapped close to each other and dissimilar samples are mapped apart from each other. In~\cite{metricsemiregression}, a metric-learning based SSR method that utilizes a Siamese neural network (SNN) with a contrastive loss function is introduced to learn a pairwise metric space. In this method, dissimilar and similar pairs, which are constructed from both labeled and unlabeled samples, are utilized for DML. To define sample similarity, the absolute differences of target values are exploited with a thresholding strategy for labeled samples. The similarity of unlabeled samples to labeled samples is estimated based on Euclidean distances between data points. Once the SNN is trained with dissimilar and similar sample pairs, the target value estimation of samples is performed on the learned metric space by the k-nearest neighbors (k-NN) algorithm. This method only considers the pairwise similarities, which may not be sufficient in accurately learning a metric space. In~\cite{Choi:2020}, an SSR method that exploits a long short-term memory (LSTM) based SNN is proposed for continuous emotion recognition. This method applies a two-stage training. In the first stage, the SNN is trained with the log-ratio loss function (which operates on sample triplets) by using only labeled samples. Then, the SNN is used to generate pseudo-labels for unlabeled samples. In the second stage, the SNN is trained with the mean squared error loss function by using pseudo-labels. This method utilizes only labeled samples for learning metric space in the first stage. This may lead to the inaccurate characterization of sample similarities in the case of the availability of small-sized labeled training sets. To address the limitations of above-mentioned methods, we introduce a novel \textbf{D}eep \textbf{M}etric \textbf{L}earning-based \textbf{S}emi-\textbf{S}upervised \textbf{R}egression (DML-S2R) method.

\section{Proposed DML-S2R Method}
Let $\mathcal{S}=\{(\boldsymbol{x}_{i}^{l},\boldsymbol{y}_{i}^{l})\}^{N}_{i = 1}$ be a set of $N$ labeled samples, where $\boldsymbol{x}_{i}^{l}$ is the $i$th labeled sample and $\boldsymbol{y}_{i}^{l}$ is its corresponding target value. Let $\mathcal{U} = \{\boldsymbol{x}_{j}^{u}\}^{M}_{j = 1}$ be the set of $M$ unlabeled samples, where $\boldsymbol{x}_{j}^{u}$ is the $j$th sample, for which the corresponding target value is unknown. The training set $\mathcal{T}$ consists of the labeled sample set $\mathcal{S}$ and the unlabeled sample set $\mathcal{U}$ ($\mathcal{T}=\mathcal{S} \cup \mathcal{U}$). We assume that $N$ is significantly lower than $M$ ($N\!<\!<\!M$). 

The proposed DML-S2R method aims to learn a metric space, in which similar samples are located close to each other, by effectively exploiting abundant unlabeled data together with scarce labeled data. This is achieved by two main steps: i) pairwise similarity modeling with scarce labeled data; and ii) triplet-based metric learning with abundant unlabeled data. For the end-to-end training of DML-S2R, we investigate an alternate learning strategy for the two steps. Fig. 1 shows a general overview of the proposed method, which is explained in detail in the following subsections.

\subsection{Pairwise Similarity Modeling with Scarce Labeled Data}
The first step aims to model the pairwise sample similarity by using a small amount of labeled samples. For this purpose, one could exploit a contrastive loss function that requires the selection of similar and dissimilar pairs based on the target values of training samples. However, in the framework of regression problems, it is challenging to define the boundaries between the target values of similar/dissimilar samples \cite{Kim:2019}. To overcome this problem, we redefine the regression problem. Instead of learning a function that estimates the target value of each sample, DML-S2R learns a function $f^d$ which estimates the target value differences of two labeled samples as:
\begin{equation} 
\label{RegressionFormulation}
f^d(\boldsymbol{x}_i^l, \boldsymbol{x}_j^l) = \boldsymbol{y}_i^l-\boldsymbol{y}_j^l.
\end{equation}
While learning to estimate target value differences, DML-S2R is also enforced to model the sample similarity for the accurate prediction of the differences. Due to this definition, instead of using $N$ samples in the first step, DML-S2R exploits $N(N\!-\!1)$ pairs of samples (i.e., all the possible pairs from the labeled sample set). This mitigates the limitations of labeled data scarcity, and thus over-fitting. For $f^d$, we utilize an SNN, which contains two identical sub-networks with shared weights. Two sub-networks of the SNN take an input pair and provide the sample features, which are concatenated to form the feature associated to the input pair. Then, this feature is fed into a fully connected layer (FC) that directly estimates the target value differences of sample pairs. Let $\boldsymbol{z}_{ij}$ be the target value difference of $\boldsymbol{x}_i^l$ and $\boldsymbol{x}_j^l$ while $i\neq j$. To learn the model parameters of $f^d$, we propose the pairwise similarity modeling loss function $\mathcal{L}_{\text{PSM}}$ as follows:
\begin{equation} 
\label{MSE}
\mathcal{L}_{\text{PSM}} = \frac{1}{N(N\!-\!1)} \sum_{i=1}^{N}\sum_{j=1}^{N} \mathds{1}_{[i \neq j]} (\boldsymbol{z}_{ij} -f^d(\boldsymbol{x}_i^l,\boldsymbol{x}_j^l))^2,
\end{equation}
where $\mathds{1}$ is the indicator function. Once the model parameters of the SNN is learnt with (2), its feature space, which encodes the pairwise sample similarity, forms the basis for the second step.

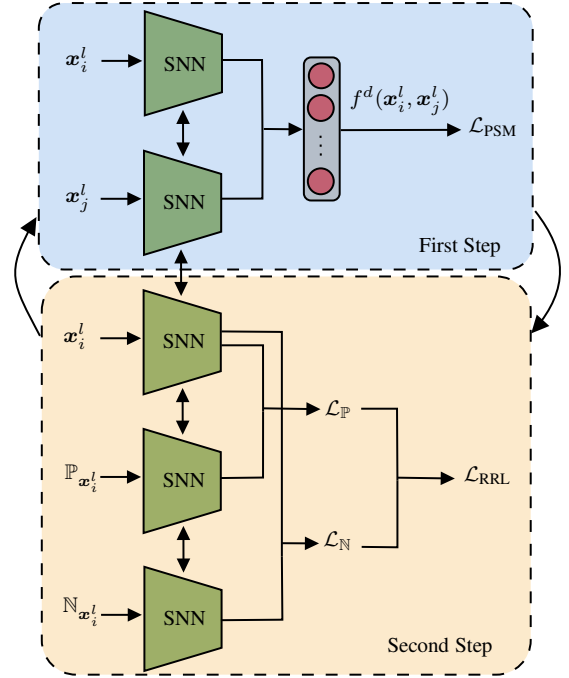
\begin{figure}[t]
    \centering

\tikzset{every picture/.style={line width=0.75pt}} 

\tikzset{every picture/.style={line width=0.75pt}} 

\begin{tikzpicture}[x=0.75pt,y=0.75pt,yscale=-1,xscale=1]

\draw  [fill={rgb, 255:red, 74; green, 144; blue, 226 }  ,fill opacity=0.26 ][dash pattern={on 4.5pt off 4.5pt}] (17.42,15.33) .. controls (17.42,8.24) and (23.16,2.5) .. (30.25,2.5) -- (253.67,2.5) .. controls (260.76,2.5) and (266.5,8.24) .. (266.5,15.33) -- (266.5,124.17) .. controls (266.5,131.26) and (260.76,137) .. (253.67,137) -- (30.25,137) .. controls (23.16,137) and (17.42,131.26) .. (17.42,124.17) -- cycle ;
\draw  [fill={rgb, 255:red, 245; green, 166; blue, 35 }  ,fill opacity=0.23 ][dash pattern={on 4.5pt off 4.5pt}] (18.92,157.83) .. controls (18.92,148.26) and (26.68,140.5) .. (36.25,140.5) -- (248.17,140.5) .. controls (257.74,140.5) and (265.5,148.26) .. (265.5,157.83) -- (265.5,324.17) .. controls (265.5,333.74) and (257.74,341.5) .. (248.17,341.5) -- (36.25,341.5) .. controls (26.68,341.5) and (18.92,333.74) .. (18.92,324.17) -- cycle ;
\draw  [fill={rgb, 255:red, 65; green, 117; blue, 5 }  ,fill opacity=0.5 ] (70.87,6.67) -- (109.8,18.35) -- (109.8,47.52) -- (70.87,59.2) -- cycle ;
\draw  [fill={rgb, 255:red, 155; green, 155; blue, 155 }  ,fill opacity=0.5 ] (151.1,33.56) .. controls (151.1,31.57) and (152.72,29.95) .. (154.71,29.95) -- (165.55,29.95) .. controls (167.54,29.95) and (169.16,31.57) .. (169.16,33.56) -- (169.16,98.55) .. controls (169.16,100.55) and (167.54,102.16) .. (165.55,102.16) -- (154.71,102.16) .. controls (152.72,102.16) and (151.1,100.55) .. (151.1,98.55) -- cycle ;
\draw  [fill={rgb, 255:red, 208; green, 2; blue, 27 }  ,fill opacity=0.5 ] (153.6,39.23) .. controls (153.6,35.76) and (156.41,32.95) .. (159.88,32.95) .. controls (163.35,32.95) and (166.16,35.76) .. (166.16,39.23) .. controls (166.16,42.7) and (163.35,45.51) .. (159.88,45.51) .. controls (156.41,45.51) and (153.6,42.7) .. (153.6,39.23) -- cycle ;
\draw  [fill={rgb, 255:red, 208; green, 2; blue, 27 }  ,fill opacity=0.5 ] (153.1,55.48) .. controls (153.1,51.87) and (156.02,48.95) .. (159.63,48.95) .. controls (163.23,48.95) and (166.16,51.87) .. (166.16,55.48) .. controls (166.16,59.08) and (163.23,62.01) .. (159.63,62.01) .. controls (156.02,62.01) and (153.1,59.08) .. (153.1,55.48) -- cycle ;
\draw  [fill={rgb, 255:red, 208; green, 2; blue, 27 }  ,fill opacity=0.5 ] (153.1,92.48) .. controls (153.1,88.87) and (156.02,85.95) .. (159.63,85.95) .. controls (163.23,85.95) and (166.16,88.87) .. (166.16,92.48) .. controls (166.16,96.08) and (163.23,99.01) .. (159.63,99.01) .. controls (156.02,99.01) and (153.1,96.08) .. (153.1,92.48) -- cycle ;
\draw  [dash pattern={on 0.84pt off 2.51pt}]  (159.51,69.45) -- (159.66,80.59) ;
\draw    (49.02,31.7) -- (66.61,31.7) ;
\draw [shift={(69.61,31.7)}, rotate = 180] [fill={rgb, 255:red, 0; green, 0; blue, 0 }  ][line width=0.08]  [draw opacity=0] (7.14,-3.43) -- (0,0) -- (7.14,3.43) -- cycle    ;
\draw    (169.68,66.65) -- (227.5,66.51) ;
\draw [shift={(230.5,66.5)}, rotate = 179.86] [fill={rgb, 255:red, 0; green, 0; blue, 0 }  ][line width=0.08]  [draw opacity=0] (7.14,-3.43) -- (0,0) -- (7.14,3.43) -- cycle    ;
\draw    (90.19,77) -- (90.15,59.4) ;
\draw [shift={(90.14,56.4)}, rotate = 89.85] [fill={rgb, 255:red, 0; green, 0; blue, 0 }  ][line width=0.08]  [draw opacity=0] (7.14,-3.43) -- (0,0) -- (7.14,3.43) -- cycle    ;
\draw [shift={(90.2,80)}, rotate = 269.85] [fill={rgb, 255:red, 0; green, 0; blue, 0 }  ][line width=0.08]  [draw opacity=0] (7.14,-3.43) -- (0,0) -- (7.14,3.43) -- cycle    ;
\draw  [fill={rgb, 255:red, 65; green, 117; blue, 5 }  ,fill opacity=0.5 ] (70.47,77.07) -- (109.4,88.75) -- (109.4,117.92) -- (70.47,129.6) -- cycle ;
\draw    (48.62,101.3) -- (66.21,101.3) ;
\draw [shift={(69.21,101.3)}, rotate = 180] [fill={rgb, 255:red, 0; green, 0; blue, 0 }  ][line width=0.08]  [draw opacity=0] (7.14,-3.43) -- (0,0) -- (7.14,3.43) -- cycle    ;
\draw    (89.19,147.4) -- (89.15,129.8) ;
\draw [shift={(89.14,126.8)}, rotate = 89.85] [fill={rgb, 255:red, 0; green, 0; blue, 0 }  ][line width=0.08]  [draw opacity=0] (7.14,-3.43) -- (0,0) -- (7.14,3.43) -- cycle    ;
\draw [shift={(89.2,150.4)}, rotate = 269.85] [fill={rgb, 255:red, 0; green, 0; blue, 0 }  ][line width=0.08]  [draw opacity=0] (7.14,-3.43) -- (0,0) -- (7.14,3.43) -- cycle    ;
\draw  [fill={rgb, 255:red, 65; green, 117; blue, 5 }  ,fill opacity=0.5 ] (70.27,147.47) -- (109.2,159.15) -- (109.2,188.32) -- (70.27,200) -- cycle ;
\draw    (48.42,171.7) -- (66.01,171.7) ;
\draw [shift={(69.01,171.7)}, rotate = 180] [fill={rgb, 255:red, 0; green, 0; blue, 0 }  ][line width=0.08]  [draw opacity=0] (7.14,-3.43) -- (0,0) -- (7.14,3.43) -- cycle    ;
\draw    (89.99,217.4) -- (89.95,199.8) ;
\draw [shift={(89.94,196.8)}, rotate = 89.85] [fill={rgb, 255:red, 0; green, 0; blue, 0 }  ][line width=0.08]  [draw opacity=0] (7.14,-3.43) -- (0,0) -- (7.14,3.43) -- cycle    ;
\draw [shift={(90,220.4)}, rotate = 269.85] [fill={rgb, 255:red, 0; green, 0; blue, 0 }  ][line width=0.08]  [draw opacity=0] (7.14,-3.43) -- (0,0) -- (7.14,3.43) -- cycle    ;
\draw  [fill={rgb, 255:red, 65; green, 117; blue, 5 }  ,fill opacity=0.5 ] (70.27,217.47) -- (109.2,229.15) -- (109.2,258.32) -- (70.27,270) -- cycle ;
\draw    (48.42,241.7) -- (66.01,241.7) ;
\draw [shift={(69.01,241.7)}, rotate = 180] [fill={rgb, 255:red, 0; green, 0; blue, 0 }  ][line width=0.08]  [draw opacity=0] (7.14,-3.43) -- (0,0) -- (7.14,3.43) -- cycle    ;
\draw    (90.39,287) -- (90.35,269.4) ;
\draw [shift={(90.34,266.4)}, rotate = 89.85] [fill={rgb, 255:red, 0; green, 0; blue, 0 }  ][line width=0.08]  [draw opacity=0] (7.14,-3.43) -- (0,0) -- (7.14,3.43) -- cycle    ;
\draw [shift={(90.4,290)}, rotate = 269.85] [fill={rgb, 255:red, 0; green, 0; blue, 0 }  ][line width=0.08]  [draw opacity=0] (7.14,-3.43) -- (0,0) -- (7.14,3.43) -- cycle    ;
\draw  [fill={rgb, 255:red, 65; green, 117; blue, 5 }  ,fill opacity=0.5 ] (70.67,287.07) -- (109.6,298.75) -- (109.6,327.92) -- (70.67,339.6) -- cycle ;
\draw    (48.82,311.3) -- (66.41,311.3) ;
\draw [shift={(69.41,311.3)}, rotate = 180] [fill={rgb, 255:red, 0; green, 0; blue, 0 }  ][line width=0.08]  [draw opacity=0] (7.14,-3.43) -- (0,0) -- (7.14,3.43) -- cycle    ;
\draw    (110,31.25) -- (130,31.25) -- (130.2,67.6) ;
\draw    (109.75,101.25) -- (130.2,101.2) -- (130.2,67.6) ;
\draw    (130.89,66.06) -- (147.5,66.22) ;
\draw [shift={(150.5,66.25)}, rotate = 180.55] [fill={rgb, 255:red, 0; green, 0; blue, 0 }  ][line width=0.08]  [draw opacity=0] (7.14,-3.43) -- (0,0) -- (7.14,3.43) -- cycle    ;
\draw    (267.75,107.5) .. controls (284.16,128.49) and (284.73,146.92) .. (267.68,167.05) ;
\draw [shift={(265.75,169.25)}, rotate = 312.14] [fill={rgb, 255:red, 0; green, 0; blue, 0 }  ][line width=0.08]  [draw opacity=0] (8.93,-4.29) -- (0,0) -- (8.93,4.29) -- cycle    ;
\draw    (109.5,175.25) -- (130.5,175.25) -- (130.53,208.6) ;
\draw    (109.25,242.25) -- (130.5,242) -- (130.53,208.6) ;
\draw    (130.47,207.06) -- (156.5,207.45) ;
\draw [shift={(159.5,207.5)}, rotate = 180.87] [fill={rgb, 255:red, 0; green, 0; blue, 0 }  ][line width=0.08]  [draw opacity=0] (7.14,-3.43) -- (0,0) -- (7.14,3.43) -- cycle    ;
\draw    (109.25,314) -- (140.25,313.75) -- (140,281.75) ;
\draw    (140.06,275.14) -- (155.75,275.23) ;
\draw [shift={(158.75,275.25)}, rotate = 180.32] [fill={rgb, 255:red, 0; green, 0; blue, 0 }  ][line width=0.08]  [draw opacity=0] (7.14,-3.43) -- (0,0) -- (7.14,3.43) -- cycle    ;
\draw    (109.5,168.25) -- (140,168.5) -- (140,281.75) ;
\draw    (180,207.25) -- (198.08,207.25) -- (198.28,243.6) ;
\draw    (180.25,277) -- (198.28,277.2) -- (198.28,243.6) ;
\draw    (198.22,242.06) -- (224.25,242.45) ;
\draw [shift={(227.25,242.5)}, rotate = 180.87] [fill={rgb, 255:red, 0; green, 0; blue, 0 }  ][line width=0.08]  [draw opacity=0] (7.14,-3.43) -- (0,0) -- (7.14,3.43) -- cycle    ;
\draw    (14.62,113.36) .. controls (0.26,138.84) and (4.01,143.96) .. (18,170.5) ;
\draw [shift={(16.25,110.5)}, rotate = 120.13] [fill={rgb, 255:red, 0; green, 0; blue, 0 }  ][line width=0.08]  [draw opacity=0] (8.93,-4.29) -- (0,0) -- (8.93,4.29) -- cycle    ;

\draw (232.7,59.03) node [anchor=north west][inner sep=0.75pt]  [font=\footnotesize]  {$\mathcal{L}_{\text{PSM}}$};
\draw (29.33,23.06) node [anchor=north west][inner sep=0.75pt]  [font=\footnotesize]  {$\boldsymbol{x}_{i}^{l}$};
\draw (76.27,27.87) node [anchor=north west][inner sep=0.75pt]  [font=\footnotesize] [align=left] {\begin{minipage}[lt]{19.95pt}\setlength\topsep{0pt}
\begin{center}
SNN
\end{center}

\end{minipage}};
\draw (172.8,43.13) node [anchor=north west][inner sep=0.75pt]  [font=\footnotesize]  {$f^{d}(\boldsymbol{x}_{i}^{l} ,\boldsymbol{x}_{j}^{l})$};
\draw (29.3,92.33) node [anchor=north west][inner sep=0.75pt]  [font=\footnotesize]  {$\boldsymbol{x}_{j}^{l}$};
\draw (29.1,233.73) node [anchor=north west][inner sep=0.75pt]  [font=\footnotesize]  {$\mathbb{P}_{\boldsymbol{x}_i^l}$};
\draw (75.87,97.47) node [anchor=north west][inner sep=0.75pt]  [font=\footnotesize] [align=left] {\begin{minipage}[lt]{19.95pt}\setlength\topsep{0pt}
\begin{center}
SNN
\end{center}

\end{minipage}};
\draw (28.3,162.73) node [anchor=north west][inner sep=0.75pt]  [font=\footnotesize]  {$\boldsymbol{x}_{i}^{l}$};
\draw (75.67,167.87) node [anchor=north west][inner sep=0.75pt]  [font=\footnotesize] [align=left] {\begin{minipage}[lt]{19.95pt}\setlength\topsep{0pt}
\begin{center}
SNN
\end{center}

\end{minipage}};
\draw (75.67,237.87) node [anchor=north west][inner sep=0.75pt]  [font=\footnotesize] [align=left] {\begin{minipage}[lt]{19.95pt}\setlength\topsep{0pt}
\begin{center}
SNN
\end{center}

\end{minipage}};
\draw (76.07,307.47) node [anchor=north west][inner sep=0.75pt]  [font=\footnotesize] [align=left] {\begin{minipage}[lt]{19.95pt}\setlength\topsep{0pt}
\begin{center}
SNN
\end{center}

\end{minipage}};
\draw (28,300.98) node [anchor=north west][inner sep=0.75pt]  [font=\footnotesize]  {$\mathbb{N}_{\boldsymbol{x}_i^l}$};
\draw (160.2,201.53) node [anchor=north west][inner sep=0.75pt]  [font=\footnotesize]  {$\mathcal{L}_{\mathbb{P}}$};
\draw (159.45,267.9) node [anchor=north west][inner sep=0.75pt]  [font=\footnotesize]  {$\mathcal{L}_{\mathbb{N}}$};
\draw (230.2,234.9) node [anchor=north west][inner sep=0.75pt]  [font=\footnotesize]  {$\mathcal{L}_{\text{RRL}}$};
\draw (195.77,119.37) node [anchor=north west][inner sep=0.75pt]  [font=\footnotesize] [align=left] {\begin{minipage}[lt]{48.97pt}\setlength\topsep{0pt}
\begin{center}
 \ First Step \ \ 
\end{center}

\end{minipage}};
\draw (185.27,321.5) node [anchor=north west][inner sep=0.75pt]  [font=\footnotesize] [align=left] {\begin{minipage}[lt]{49.45pt}\setlength\topsep{0pt}
\begin{center}
Second Step
\end{center}

\end{minipage}};

\end{tikzpicture}

    \caption{Illustration of the proposed DML-S2R method.}
\label{fig:model}
\end{figure}
\subsection{Triplet-Based Metric Learning with Abundant Unlabeled Data}
The second step aims to learn a metric space (where similar samples are located close to each other) when the number of labeled samples is limited. This is achieved by triplet-based DML that takes into account not only labeled samples but also unlabeled samples. Triplet-based DML operates on triplets for the characterization of a metric space. Accordingly, we convert the SNN of the first step to the triplet-based SNN by replicating one of its identical sub-networks three times without changing weights, and thus make it appropriate for triplet-based DML. A standard triplet consists of a sample anchor and a positive sample (which is similar to the anchor) and a negative sample (which is dissimilar to the anchor). For the construction of triplets, anchor samples are selected from the labeled set $\mathcal{S}$ and positive and negative samples are selected from the unlabeled set $\mathcal{U}$. To effectively exploit abundant unlabeled data, we create a set of positive and negative samples per anchor. This leads to a faster convergence compared to having only one positive and one negative sample per anchor \cite{rankedlistloss}. For an anchor $\boldsymbol{x}_a^l \in \mathcal{S}$, the set of $k$ positive samples $\mathbb{P}_{\boldsymbol{x}_a^l}$ and the set of $k$ negative samples $\mathbb{N}_{\boldsymbol{x}_a^l}$ are selected with $f^d$ from the first step based on their estimated target value differences with the anchor. In detail, all the possible pairs between the anchor $\boldsymbol{x}_a^l$ and each unlabeled sample $\boldsymbol{x}_i^u \in \mathcal{U}$ are created. Then, the target value difference of each pair is estimated by using $f^d$ from the first step. The $k$ unlabeled samples having the smallest difference with the anchor are selected for $\mathbb{P}_{\boldsymbol{x}_a^l}$, while the $k$ unlabeled samples having the highest difference with the anchor are selected for $\mathbb{N}_{\boldsymbol{x}_a^l}$. The positive-negative set selection procedure is shown in Algorithm 1. After the selection of the positive and negative sets for all anchor samples, we employ the ranked list loss function \cite{rankedlistloss} as follows:
\begin{equation}
\begin{aligned}
w(\boldsymbol{x}_a^l,\boldsymbol{x}_j^u) &= \text{exp}(\tau(d(\boldsymbol{x}_a^l,\boldsymbol{x}_j^u)-(\alpha-m))),\\
\mathcal{L}_{\mathbb{S}}(\boldsymbol{x}_a^l, \mathbb{S}) &= \sum_{\boldsymbol{x}_j^u \in \mathbb{S}} \frac{w(\boldsymbol{x}_a^l,\boldsymbol{x}_j^u)}{\sum_{\boldsymbol{x}_j^u \in \mathbb{S}} w(\boldsymbol{x}_a^l,\boldsymbol{x}_j^u)} \mathcal{L}_m(\boldsymbol{x}_a^l, \boldsymbol{x}_j^u),\\
\mathcal{L}_{\text{RLL}} &= \frac{1}{2N}\sum_{i=1}^N\mathcal{L}_{\mathbb{P}}(\boldsymbol{x}_a^l, \mathbb{P}_{\boldsymbol{x}_a^l}) + \mathcal{L}_{\mathbb{N}}(\boldsymbol{x}_a^l, \mathbb{N}_{\boldsymbol{x}_a^l}),
\end{aligned}
\end{equation}
where $\tau$ is the temperature parameter, $\alpha$ is the negative sample boundary, $m$ is the margin parameter, $d$ measures the Euclidean distance between two samples in the feature space and $\mathcal{L}_m$ is the margin loss function. $\mathcal{L}_{\text{RLL}}$ pulls a set of positive samples closer than the set of negatives by the margin $m$ on the feature space of the SNN. 
\SetKwComment{Comment}{}{}
\begin{algorithm}[t]
\SetAlgoLined
\SetKwFunction{Union}{Union}\SetKwFunction{FindCompress}{FindCompress}
\SetKwInOut{Input}{Input}\SetKwInOut{Output}{Output}
\Input{$\boldsymbol{x}_a^l, \mathcal{U}, f^d, k$}
\Output{$\mathbb{P}_{\boldsymbol{x}_a^l}, \mathbb{N}_{\boldsymbol{x}_a^l}$}

\BlankLine

$\mathbb{P}_{\boldsymbol{x}_a^l} = \emptyset, \mathbb{N}_{\boldsymbol{x}_a^l}=\emptyset$

\While{$|\mathbb{P}_{\boldsymbol{x}_a^l}| \leq k$}{
$\argmin\limits_{\boldsymbol{x}_i^u\in\mathcal{U} \setminus \mathbb{P}_{\boldsymbol{x}_a^l}} f^d(\boldsymbol{x}_a^l, \boldsymbol{x}_i^u) \to \mathbb{P}_{\boldsymbol{x}_a^l}$ {\centering \footnotesize (Positive set selection)}
}

\While{$|\mathbb{N}_{\boldsymbol{x}_a^l}| \leq k$}{
$\argmax\limits_{\boldsymbol{x}_i^u\in\mathcal{U} \setminus \mathbb{N}_{\boldsymbol{x}_a^l}} f^d(\boldsymbol{x}_a^l, \boldsymbol{x}_i^u) \to \mathbb{N}_{\boldsymbol{x}_a^l}$ {\footnotesize (Negative set selection)}
}
\KwRet{$\mathbb{P}_{\boldsymbol{x}_a^l}$, $\mathbb{N}_{\boldsymbol{x}_a^l}$} 
\label{TripletSelectionVariant1}
\caption{Positive-negative set selection for an anchor within the proposed DML-S2R method}
\end{algorithm}

It is worth noting that the effectiveness of each step in DML-S2R depends on each other. Inaccurate learning of $f^d$ in the first step leads to incorrect selection of positive-negative sets in the second step. If the metric space is not accurately learned in the second step, the weights of the SNN can not be effectively learned in the first step due to a small number of labeled samples. This prevents to utilize standard joint learning strategies for DML-S2R. To accurately learn both steps, we investigate an alternate learning strategy, in which the SNN is trained for both steps within the consecutive training epochs. In detail, the whole learning procedure starts with training the SNN for one epoch by the first step while minimizing $\mathcal{L}_{\text{PSM}}$. Then, the SNN is trained for one epoch by the second step with all the anchor samples and the associated positive-negative sets while minimizing $\mathcal{L}_{\text{RRL}}$. Training continues while alternating between the two steps until convergence of both loss functions. Once the training of DML-S2R is completed, the target value estimation of a new sample is achieved based on $f^d$ as follows: 
\begin{equation} 
\label{PredictionFormulation}
 \boldsymbol{y}^* = \frac{1}{N} \sum_{i=1}^{N} \left( \frac{f^d(\boldsymbol{x}^*, \boldsymbol{x}_i^l) - f^d(\boldsymbol{x}_i^l, \boldsymbol{x}^*)}{2}+ \boldsymbol{y}_i^l \right).
\end{equation}
\section{Experimental Results}
Experiments were conducted on Boston Housing \cite{HousingPrices}, Superconductivity \cite{Superconductivity} and Air Quality \cite{Airquality} datasets associated to the regression problems of housing value estimation, critical temperature estimation of a superconductor and benzene estimation for pollution monitoring, respectively. The Boston Housing dataset includes 506 samples, each of which is associated with 13 variables. The Superconductivity dataset includes 21263 samples, while each sample is associated with 81 variables. The Air Quality dataset consists of 9358 samples, each of which is associated with 14 variables. For the construction of the unlabeled set, we randomly selected 1000 samples for the Superconductivity and Air Quality datasets, while 200 samples were randomly chosen for the Boston Housing dataset. In the experiments, the number of labeled samples is varied as $|S|=10,20,50$ for each dataset. The rest of the samples were used as the test set for each dataset. All the samples from three datasets were normalized by using the min-max normalization approach. In the experiments, two hidden layers, each of which includes 100 neurons, were used for the SNN of DML-S2R. The parameter $k$ was set to 5. We trained our method for 30 epochs by using the Adam optimizer with the initial learning rate of 0.001. The results are provided in terms of mean absolute error (MAE). To perform the ablation study of the proposed DML-S2R method, we compared it with only using the first step of DML-S2R. To assess the effectiveness of the proposed DML-S2R method, we compared it with the cotraining-style SSR algorithm (COREG)~\cite{COREG} and metric-based SSR (denoted as MSSR) method~\cite{metricsemiregression}. 
\begin{table}[t] 
\centering
\renewcommand{\arraystretch}{0.9}
\caption{Mean absolute error (MAE) scores when the different steps of the proposed DML-S2R method were utilized (Superconductivity dataset).}
\label{tab:Ablation}
\begin{tabular}{
@{\hskip -0.025in}*{2}{>{\centering\arraybackslash}p{0.08\textwidth}}*{4}{>{\centering\arraybackslash}p{0.05\textwidth}}@{\hskip -0.01in}}
\toprule
\multicolumn{2}{c}{\textbf{Steps of DML-S2R}} & \multicolumn{4}{c}{\textbf{Number of Labeled Samples}} \tabularnewline 
\cmidrule(lr{1em}){1-2}\cmidrule(lr{1em}){3-6}
1\textsuperscript{st}& 2\textsuperscript{nd} & 10 & 20 & 50 & 100\tabularnewline
\toprule
\cmark & \xmark &  23.5 & 22.8 & 18.2 & 17.2 \tabularnewline
\cmark & \cmark &  \textbf{22.3} & \textbf{18.2} & \textbf{16.2} & \textbf{15.8} \tabularnewline
\bottomrule
\end{tabular}
\end{table}

\begin{table}[t] 
\centering
\renewcommand{\arraystretch}{0.9}
\caption{Mean absolute error (MAE) scores obtained by COREG, MSSR and the proposed DML-S2R method (Superconductivity dataset).}
\label{tab:Superconductivity}
\begin{tabular}{
@{}*{1}{>{\raggedright\arraybackslash}p{0.16\textwidth}}*{3}{>{\centering\arraybackslash}p{0.07\textwidth}}@{\hskip -0.01in}}
\toprule
\multirow{3}{0.23\textwidth}[2pt]{\raggedright \textbf{Method}} & \multicolumn{3}{c}{\textbf{Number of Labeled Samples}} \tabularnewline 
\cmidrule(lr{1em}){2-4}
 & 10 & 20 & 50 \tabularnewline
\toprule
COREG~\cite{COREG} & 23.1 & 20.5 & 17.2 \tabularnewline
MSSR~\cite{metricsemiregression} & 22.4 & 21.2 & 18.1 \tabularnewline
DML-S2R (Ours) & \textbf{22.3} & \textbf{18.2} & \textbf{16.2} \tabularnewline
\bottomrule
\end{tabular}
\end{table}
\begin{table}[t] 
\centering
\renewcommand{\arraystretch}{0.9}
\caption{Mean absolute error (MAE) scores obtained by COREG, MSSR and the proposed DML-S2R method (Boston Housing dataset).}
\label{tab:Boston}
\begin{tabular}{
@{}*{1}{>{\raggedright\arraybackslash}p{0.16\textwidth}}*{3}{>{\centering\arraybackslash}p{0.07\textwidth}}@{\hskip -0.01in}}
\toprule
\multirow{3}{0.23\textwidth}[2pt]{\raggedright \textbf{Method}} & \multicolumn{3}{c}{\textbf{Number of Labeled Samples}} \tabularnewline 
\cmidrule(lr{1em}){2-4}
 & 10 & 20 & 50 \tabularnewline
\toprule
COREG~\cite{COREG} & 8.2 & 7.9 & 5.1 \tabularnewline
MSSR~\cite{metricsemiregression} & 6.5 & 5.6 & 4.9 \tabularnewline
DML-S2R (Ours) & \textbf{6.1} & \textbf{5.4} & \textbf{4.5} \tabularnewline
\bottomrule
\end{tabular}
\end{table}
\begin{table}[t] 
\centering
\renewcommand{\arraystretch}{0.9}
\caption{Mean absolute error (MAE) scores obtained by COREG, MSSR and the proposed DML-S2R method (Air Quality dataset).}
\label{tab:Airquality}
\begin{tabular}{
@{}*{1}{>{\raggedright\arraybackslash}p{0.16\textwidth}}*{3}{>{\centering\arraybackslash}p{0.07\textwidth}}@{\hskip -0.01in}}
\toprule
\multirow{3}{0.23\textwidth}[2pt]{\raggedright \textbf{Method}} & \multicolumn{3}{c}{\textbf{Number of Labeled Samples}} \tabularnewline 
\cmidrule(lr{1em}){2-4}
 & 10 & 20 & 50 \tabularnewline
\toprule
COREG~\cite{COREG} & 12.4 & 11.9 & 9.5 \tabularnewline
MSSR~\cite{metricsemiregression} & 17.9 & 17.3 & 12.7 \tabularnewline
DML-S2R (Ours) & \textbf{10.9} & \textbf{6.0} & \textbf{3.3} \tabularnewline
\bottomrule
\end{tabular}
\end{table}

Table \ref{tab:Ablation} shows the regression performances obtained on the Superconductivity dataset when: i) only the first step of the proposed DML-S2R method is applied; and ii) both steps of DML-S2R are applied. By analyzing the table, one can see that utilizing both steps of DML-S2R results in significantly higher accuracies than using only the first step of DML-S2R under different numbers of labeled samples. As an example, DML-S2R achieves an MAE of 18.2 with 20 labeled samples, whereas using only the first step of DML-S2R results in an MAE of 22.8 with the same number of labeled samples. This is due to the fact that in the second step of DML-S2R, unlabeled samples are accurately exploited to learn a deep metric space that leads to a more accurate target value estimation. This also shows the success of the alternate learning strategy for learning both steps. We observed similar behaviours for the other considered datasets (not reported for space constraints). Tables \ref{tab:Superconductivity}-\ref{tab:Airquality} show the regression performances of COREG, MSSR and proposed DML-S2R with the different numbers of labeled samples for all the datasets. One can see from the tables that our method provides the highest accuracies for all datasets. As an example, DML-S2R achieves an MAE of 3.3 with 50 labeled samples, whereas COREG provides an MAE of 9.5 and MSSR achieves an MAE of 12.7 with the same number of labeled samples for the Superconductivity dataset. This is due to the fact that our method effectively models the pairwise similarity of samples by using scarce labeled data, while accurately learning a deep metric space by exploiting labeled and unlabeled samples together. 
\section{Conclusion}
In this paper, we have presented a novel deep metric learning-based semi-supervised regression (DML-S2R) method for parameter estimation problems. DML-S2R includes two consecutive steps. The first step aims at modeling the pairwise similarities of samples when the labeled data is scarce. This is achieved by learning to estimate the target value differences of the labeled sample pairs with an SNN. Due to this step, the proposed DML-S2R method overcomes the challenges of defining pairwise similar/dissimilar samples based on the target values of labeled samples in the framework of regression problems. The second step aims at learning a metric space that utilizes not only labeled samples but also unlabeled samples to further enrich modeling sample similarities with abundant unlabeled data. This is achieved by employing the SNN of the first step for triplet-based DML where positive-negative samples in each triplet are selected from unlabeled samples based on the SNN. For the whole learning procedure of DML-S2R, we investigate an alternate learning strategy, in which the SNN is trained for both steps through consecutive training epochs. Due to this strategy, the encoded information by the SNN in each step becomes a guidance for learning phase of the other step. Experimental results show the effectiveness of proposed DML-S2R compared to state-of-the-art SSR methods. It is worth noting that DML-S2R is independent from the type of the SNN architecture, and thus suitable to be integrated into any Siamese-based deep architecture. As a future work, we plan to apply DML-S2R to the parameter estimation problems in image domain (e.g., age prediction, emotion recognition, etc.).
\section{Acknowledgments}
This work is funded by the European Research Council (ERC) through the ERC-2017-STG BigEarth Project under Grant 759764.
\vfill
\pagebreak
\bibliographystyle{IEEEbib}
\bibliography{references}
\end{document}